\documentclass[10pt,journal,compsoc]{IEEEtran}
\usepackage{cite}
\usepackage[pdftex]{graphicx}
\usepackage{amsmath}
\usepackage{amsfonts}
\usepackage{xcolor}

\usepackage{color}
\usepackage[symbol]{footmisc}

\begin{document}

    \title{Interpretation of Time-Series Deep Models: \\ A Survey}

    \author{
    Ziqi~Zhao{$^\ast$},
    Yucheng~Shi{$^\ast$},
    Shushan~Wu{$^\ast$},
    Fan~Yang,
    Wenzhan~Song,
    Ninghao~Liu
    \IEEEcompsocitemizethanks{
    \IEEEcompsocthanksitem {$^\ast$} These authors contributed equally to this paper.
\IEEEcompsocthanksitem Yucheng~Shi and Ninghao~Liu are with the School of Computing, University of Georgia, GA, USA. E-mail: \{yucheng.shi, ninghao.liu\}@uga.edu.
\IEEEcompsocthanksitem Ziqi Zhao is with the Department of Computer Science and Engineering, Texas A\&M University, TX, USA. E-mail: astrajoan@tamu.edu.
\IEEEcompsocthanksitem Shushan Wu is with the Department of Statistics, University of Georgia, GA, USA. E-mail: shushan.wu@uga.edu.
\IEEEcompsocthanksitem Fan Yang is with the Department of Computer Science, Rice University, TX, USA. E-mail: fyang@rice.edu.
\IEEEcompsocthanksitem Wenzhan Song is with the School of Electrical and Computer Engineering, University of Georgia, GA, USA. E-mail: wsong@uga.edu.
}
    }
    
    \markboth{Preprint}
    {Interpretation of Time-Series Deep Models: A Survey}
    
    \IEEEtitleabstractindextext{
        \begin{abstract}
            Deep learning models developed for time-series associated tasks have become more widely researched nowadays. However, due to the unintuitive nature of time-series data, the interpretability problem -- where we understand what is under the hood of these models -- becomes crucial. The advancement of similar studies in computer vision has given rise to many post-hoc methods, which can also shed light on how to explain time-series models. In this paper, we present a wide range of post-hoc interpretation methods for time-series models based on backpropagation, perturbation, and approximation. We also want to bring focus onto inherently interpretable models, a novel category of interpretation where human-understandable information is designed within the models. Furthermore, we introduce some common evaluation metrics used for the explanations, and propose several directions of future researches on the time-series interpretability problem. As a highlight, our work summarizes not only the well-established interpretation methods, but also a handful of fairly recent and under-developed techniques, which we hope to capture their essence and spark future endeavours to innovate and improvise.
        \end{abstract}
        
        \begin{IEEEkeywords}
            Time-series models, interpretability, post-hoc interpretation, interpretable models.
        \end{IEEEkeywords}
    }
    
    \maketitle
    \IEEEdisplaynontitleabstractindextext
    \IEEEpeerreviewmaketitle

    \IEEEraisesectionheading{\section{Introduction}\label{sec:introduction}}

    \IEEEPARstart{T}{ime} series data refers to data that is collected and recorded sequentially over time from diverse sources such as healthcare~\cite{clemente2020helena}, finance~\cite{idrees2019prediction}, economics~\cite{marcellino2007comparison}, energy~\cite{zhen2021photovoltaic}, and climate~\cite{karevan2020transductive}. Various machine learning models, including convolutional neural networks (CNNs)~\cite{he2016deep}, recurrent neural networks (RNNs)~\cite{hochreiter1997long}, and Transformers~\cite{wen2022transformers}, have been used to process time series data for tasks such as classification~\cite{wang2017time}, forecasting~\cite{schockaert2020attention}, regression~\cite{ostrom1990time}, and generation~\cite{li2021learning}. However, many of these deep models lack interpretability, as they are developed prioritizing performance instead of transparency, making them black-box models.   
    
    The lack of interpretability in models can pose challenges in critical applications where trust, fairness, privacy, and safety are paramount. The reliability of predictions made by time series models is crucial in many important decision-making contexts. For example, in finance, model predicted stock prices are used to make investment decisions~\cite{idrees2019prediction}. In weather forecasting, predictions of storms are used to make decisions about evacuations and other emergency measures~\cite{karevan2020transductive}. Thus, ensuring the interpretability of time series models is crucial in real-world applications. 
    
    To tackle the above issues, researchers have proposed various explanation methods for the interpretation of time series models. Generally speaking, there are mainly two kinds of interpretation. For the first kind, they interpret those black-box models, e.g., the recurrent neural network~\cite{rumelhart1986learning}. We name them post-hoc interpretation methods because this kind of method explains the time series models after the model inference. The post-hoc interpretation methods mainly include backpropagation-based methods, e.g., saliency map~\cite{simonyan2013deep}, perturbation-based methods, e.g., ConvTimeNet~\cite{DBLP:journals/corr/abs-1904-12546}, and approximation-based methods, e.g., LIMEsegment~\cite{sivill2022limesegment}. However, the above interpreting process is decoupled from the model training process, which requires extra computing resources and the user can not access the explanation result until the inference is finished. To free from this flaw, researchers propose a second kind of interpretation, which is extracted from inherently interpretable models in the process of decision-making or while being trained, e.g, attention-based models~\cite{bahdanau2014neural}. 
    
    Faced with so many rising time series explanation methods, a comprehensive examination of the methods and evaluations of techniques for interpreting deep learning models on time series is necessary. The existing surveys mainly focus on the explanation methods for image, text, and graph domains~\cite{buhrmester2021analysis, lertvittayakumjorn2021explanation, lyu2022towards, yuan2020explainability}. Surveys for the explainability of time series deep learning models remain scarce. 

    To bridge this gap, we take the initial step to give a systematic study of time-series model explanation methods. Initially, we provide an overview of existing Time Series Models, encompassing Convolutional Neural Network (CNN) based, Recurrent Neural Network (RNN) based, Transformer based, and Graph Neural Network (GNN) based models. Then, we introduce two types of time-series interpretation methods in detail, namely post-hoc interpretation methods and inherently interpretable models. Finally, we summarize widely utilized evaluation metrics for time series interpretation, and discuss the limitations of existing methods and future research directions. The contributions of this work are summarized as below:
    \begin{itemize}
        \item This survey paper presents a thorough examination of current methods for explaining deep time-series models. To our knowledge, this is the first and only survey dedicated solely to this subject.
        \item We provide a comprehensive overview of the latest advancements in time-series deep models and their diverse architectures.
        \item We explore the diverse methods for interpreting the outcomes of time-series deep models in this survey. A meticulous examination of the methodology, advantages, disadvantages, and comparative analysis of the different methods is provided. 
        \item We outline the existing methods limitations and future directions in the interpretation of time-series deep models, offering valuable insight for the development of new methods on time series model interpretation. 
    \end{itemize}

    \textbf{Paper selection.} The papers included in this survey are selected from a wide range of top venues between 2014 and 2023, in the areas of machine learning (ICLR, ICML, NeurIPS), artificial intelligence (AAAI, WWW), natural language processing (EMNLP), data mining (ICDM, SIGKDD), and signal processing (ICASSP). Besides the accepted papers, we also include papers in the arXiv online database, either based on their high numbers of citations, or due to the novelty of knowledge presented in some latest works not yet formally represented in publications. We review papers mainly based on content-wise metrics, i.e. their correlation with time series models, the amount of focus on explaining model decisions, and the contributions of such works towards the evolution of interpretability methods.
    
    \textbf{Related surveys and differences to this survey.}
    The previous works~\cite{zhang2018visual,belinkov2020interpretability} mainly focus on interpretation models on specific domains other than time series. Meanwhile, existing work in the time series domain either focused on benchmarking the interpreting models in time series~\cite{ismail2020benchmarking}, or only discussed explanation methods for a certain time-series task such as classification~\cite{theissler2022explainable}, or mainly focused on post-hoc explanation methods~\cite{rojat2021explainable}, while overlooking other types of interpretation methods. In contrast, our work differs from theirs as we discuss more general types of explanation methods that can be applied to various kinds of tasks. Specifically, in addition to discussing traditional post-hoc models, we also cover recent inherently interpretable models. Moreover, we provide comprehensive reviews of current time series models and evaluation metrics, making this survey self-contained.

    \section{Time-Series Deep Models}\label{sec:models}
    
   Machine learning models, such as CNN, RNN and Transformer, excel at capturing structure and patterns in time series data. Currently, they have been widely applied in analyzing time series data. In the following section, we provide a brief overview to these models.
    
    \begin{figure*}[htbp]
    \centering
    \includegraphics[width=\linewidth]{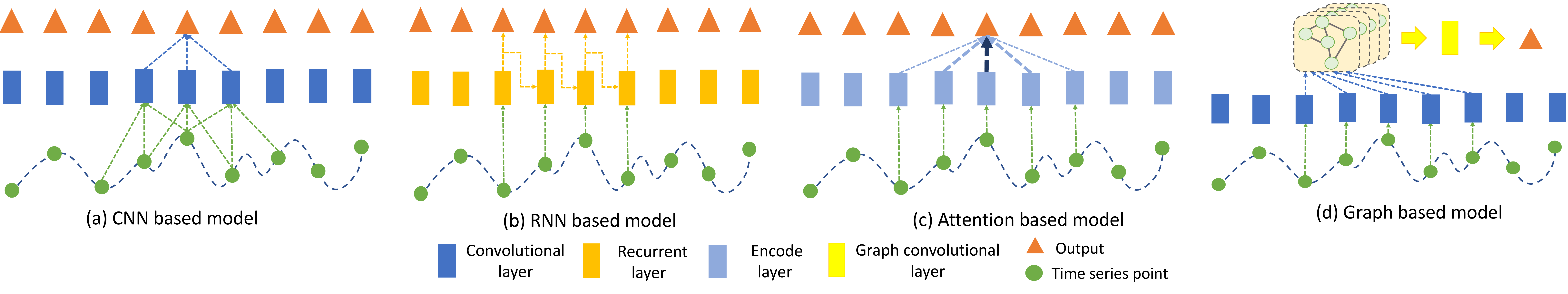}
    \caption{The architecture visualization for four types of time Series models}
    \vspace{-10pt}
    \end{figure*}
    
    \subsection{CNN-based Models}
    
    Convolutional networks have demonstrated significant capability in localized feature extraction, which has motivated researchers to adopt them in the time series domain. We will introduce three types of CNN in this section: (1) vanilla convolutional neural networks (Section~\ref{CNN}), (2) fully convolutional networks (Section~\ref{fcn}), and (3) temporal convolutional networks (Section~\ref{TCN}).
    
    \subsubsection{Convolutional Neural Networks}
    \label{CNN}
    A general form of CNN applying to a time series $\mathbf{x} = \{x_1, x_2, ..., x_n\}$ at timestamp $t$ can be defined as:
    \begin{equation}
        \label{conv}
        \mathbf{h} = \sigma(W^C * \mathbf{x}_{t-\frac{l}{2}:t+\frac{l}{2}}+b),
    \end{equation}
     where $\mathbf{h}$ denotes the obtained embeddings, $*$ denotes the element-wise product, $W^C$ stands for the learnable kernel with length $l$, $b$ stands for bias, and $\sigma$ denotes the activation function. 
     
    Some representative works include the MC-DCNN~\cite{zheng2016exploiting}, where a multi-channel deep CNN model is proposed for multivariate time series classification, and the ROCKET~\cite{dempster2020rocket} where the employment of random convolutional kernels achieves state-of-the-art performance in the time series classification task. 
    
    
    \subsubsection{Fully Convolutional Networks}
    \label{fcn}
    Fully Convolutional Networks (FCN) is originally designed for image segmentation tasks~\cite{long2015fully}. Wang et al. first adopt FCN in univariate time series classification~\cite{wang2017time}. In their design, the basic convolution block $\texttt{Block()}$ is defined as:
    \begin{equation}
        \label{convblock}
        \begin{aligned}
        \mathbf{h}_{b1}&=\sigma(W^C * \mathbf{x}_{t-\frac{l}{2}:t+\frac{l}{2}}+b) \\
        \mathbf{h}_{b2}&=\operatorname{BN}(\mathbf{h}_{b1}) \\
        \mathbf{h}&=\operatorname{ReLU}(\mathbf{h}_{b2})
        \end{aligned},
    \end{equation}
    where BN and ReLU represent the batch normalization operation and the ReLU activation function, respectively. And $\mathbf{h}_{b1}$ and $\mathbf{h}_{b2}$ denote the intermediate embeddings within the convolution block. Multiple convolution blocks can be stacked with different kernel sizes for a deeper model. After the convolution blocks, the features will be down-sampled with a global average pooling layer.
    
    Inspired by the success of ResNet in CV domain~\cite{he2016deep}, a similar design with residual connection for time series classification is proposed in~\cite{wang2017time}. Denoted the convolutional block with the number of filters $k$ in Equation~\ref{convblock} by Block$_k$, the residual block can be defined as
    \begin{equation}
        \label{resblock}
        \begin{aligned}
        \mathbf{h}_{1}&=\operatorname{Block}_{k_{1}}(\mathbf{x}_{t-\frac{l}{2}:t+\frac{l}{2}}) \\
        \mathbf{h}_{2}&=\operatorname{Block}_{k_{2}}\left(\mathbf{h}_{1}\right) \\
        \mathbf{h}_{3}&= \operatorname{Block}_{k_{3}}\left(\mathbf{h}_{2}\right) \\
        \hat{\mathbf{h}}_{3}&=\mathbf{h}_{3}+\mathbf{x}_{t-\frac{l}{2}:t+\frac{l}{2}} \\
        \mathbf{h}&=\operatorname{ReLU}(\hat{\mathbf{h}}_{3})
        \end{aligned},
    \end{equation}
    where we choose three convolution blocks to form the residual block following the same settings in~\cite{wang2017time}. It should be noted that the number and type of convolution blocks can be changed accordingly based on specific tasks. A ResNet-like design will stack multiple residual blocks together for a much deeper network. Before the final output, the features will be passed into a global average pooling layer and a softmax layer.

    \subsubsection{Temporal Convolutional Networks}
    \label{TCN}
    Temporal Convolutional Network (TCN) is a convolutional architecture designed for sequence modeling~\cite{bai2018empirical}. The TCN can be viewed as a combination of 1-dimensional FCN and causal convolutions. The FCN design will ensure the TCN can produce an output with the same size as the input. The causal convolutions will ensure that future information not be leaked into the past. To avoid an extremely deep network while achieving a large receptive field, the dilated convolutions~\cite{oord2016wavenet} are utilized, which denoted as:
    \begin{equation}
        \label{tcn}
        \mathbf{h} = \sigma(W^D \cdot \mathbf{x}_{\{t-d \cdot i\} :\{ t + d \cdot i\}} + b),
    \end{equation}
    where $d$ denotes the dilation factor, and $W^D$ is the dilated learnable kernel. $\mathbf{x}_{\{t-d \cdot i\} :\{ t + d \cdot i\}}$ is the dilated time series, where every $d$ values is skipped before inputting to network. 

    \subsection{RNN-based Models}
    Recurrent neural networks (RNNs) are a type of neural network designed for processing sequential data, such as time series data. The concept of RNNs was first introduced by David Rumelhart in the 1980s (Section~\ref{rnn})\cite{rumelhart1986learning}. A well-known variation of RNNs is the long short-term memory (LSTM) network, proposed by Hochreiter and Schmidhuber in the 1990s (Section~\ref{lstm})~\cite{hochreiter1997long}. Another important RNN-based model is the Gated Recurrent Unit (GRU), which is a modification of the LSTM and was proposed by Cho et al. (Section~\ref{gru})~\cite{cho2014learning}.
    
    \subsubsection{Recurrent Neural Networks}
    \label{rnn}
    Elman Networks~\cite{rumelhart1986learning}, also known as Elman Recurrent Neural Networks (RNNs), are a type of recurrent neural network architecture that is commonly used for processing time series data. Elman Networks have a simple and straightforward structure that includes recurrent connections, allowing them to capture temporal dependencies in the data. The basic definition of an Elman Network can be summarized as
    \begin{equation}
        h_t=\sigma\left(W^{In} x_{t}+W^{Rec} h_{t-1}+b\right),
    \end{equation}
    where we denote $h_{t}$ as the hidden state at time step $t$, $x_{t}$ as the current input, $h_{t-1}$ as the previous activation values saved in the context units, $W^{In}$ as the weight matrix for the input connections, $W^{Rec}$ as the weight matrix for the recurrent connections, $b$ as the bias, and $\sigma$ as the activation function.

    \subsubsection{Long Short-Term Memory}
    \label{lstm}
    The long short-term memory (LSTM) model is proposed to address the vanishing gradient problem in the vanilla RNN model~\cite{hochreiter1997long}. LSTM utilizes a special cell unit called 'Gate' to save important information while deleting irrelevant information. There is a total of three kinds of gates: an input gate $G^{i}$, an output gate $G^{o}$, and a forget gate $G^{f}$. The update gate $G^{u}$ controls how much information from the cell input activation vector $\tilde{c}_{t}$ will be used to update the cell state vector $c_{t}$. The forget gate $G^{f}$ decides whether the past information should be deleted or kept. And the output gate $G^{o}$ will decide how the learning information affects the final output. The model is indicated by
    \begin{equation}
    \begin{aligned}
    G^{i} &= \sigma(W^{xi}x_t + W^{hi}h_{t-1} + b_i) \\
    G^{o} &= \sigma(W^{xo}x_t + W^{ho}h_{t-1} + b_o) \\
    G^{f} &= \sigma(W^{xf}x_t + W^{hf}h_{t-1} + b_f) \\
    c'_t &= \sigma(W^{xc}x_t + W^{hc}h_{t-1} + b_c) \\
    c_t &= G^{i} \times c'_t + G^{f} \times c_{t-1} \\
    h_t &= G^{o} \times \sigma(c_t) \\
    \end{aligned}
    \end{equation}
    where $W^{xi}$, $W^{hi}$, $W^{xo}$, $W^{ho}$, $W^{xf}$, $W^{hf}$, $W^{xc}$, $W^{hc}$ and $b_i$, $b_o$, $b_f$, $b_c$ are the trainable parameters that control the gates $G^{i}$, $G^{f}$, $G^{o}$, and cell state $c'_t$ respectively. $\sigma$ is the activation function.
    
    \subsubsection{Gated Recurrent Unit}
    \label{gru}
    Though proven effective, the LSTM network has a high computational cost problem. As a simplified version of LSTM, Gated recurrent unit (GRU) has similar cell units like forget gate but without the output gate, resulting in fewer parameters~\cite{cho2014learning}. The model is expressed as
    \begin{equation}
        \begin{array}{l}
        G^{u}= \sigma(W^{xu}x_t + W^{hu}h_{t-1} + b_u) \\
        G^{r}=\sigma(W^{xr}x_t + W^{hr}h_{t-1} + b_r)\\
        h'_t = \sigma(W^{xh}x_t + W^{hh}(G^{r} \odot h_{t-1}) + b_h)\\
        h_{t}=G^{u} \times h'_{t}+\left(1-G^{u}\right) \times h_{t-1} \\
        \end{array},
    \end{equation}
    where $W^{xu}$, $W^{hu}$, $W^{xr}$, $W^{hr}$, $b_{u}$ and $b_{r}$ are the trainable parameters (weights and bias) that control the gates  $G^{u}$ and $G^{r}$, respectively. As shown above, GRU does not have a separate cell state like LSTM. Instead, GRU directly updates the hidden state using the reset and update gates.
    
    \subsection{Transformer-based Models}
    Transformer~\cite{vaswani2017attention} has gained tremendous attention for its superior performance in the natural language processing~\cite{devlin2018bert} and computer vision~\cite{dosovitskiy2020image}. The transformer model architecture relies solely on the attention mechanism to extract the global dependencies between input and output sequences. Inspired by its strong modeling ability on long-range dependencies in sequential data, various types of research have been proposed to apply transformers in time series domain~\cite{wen2022transformers}.
    
     The vanilla transformer model consists of a stack of encoders modules (six modules in~\cite{vaswani2017attention}) and the same number of decoders modules. Each encoder module comprises two sub-layers, including a multi-attention layer and a feed-forward neural network (FFNN) layer. In comparison, the decoder block contains three sub-layers, including two multi-attention layers and one FFNN layer.
     
    The core design of the transformer is the multi-head attention layer. The vanilla attention operation can be defined as~\cite{vaswani2017attention}:
    \begin{equation}
        \operatorname{Attention}(Q, K, V)=\operatorname{softmax}\left(\frac{Q K^{T}}{\sqrt{D_{k}}}\right)V,
    \end{equation}
    where $Q \in R^{N \times D_{k}}$, $K\in R^{M \times D_{k}}$, and $V\in R^{M \times D_{v}}$ are three matrices corresponding to the queries, the keys, and the values. Here $N$, $M$ denote the instances of a number of queries and keys (or values), $D_k$, $D_v$ denote the channel dimensions of keys (or queries) and values. Then the vanilla multi-head attention is defined as~\cite{vaswani2017attention}:
    \begin{equation}
    \begin{array}{l}
        \text {head}_{i}=\operatorname{Attention }\left(QW_{i}^{Q}, KW_{i}^{K}, V W_{i}^{V}\right),\\
        \operatorname{MultiheadAttention}(Q, K, V)\\=\operatorname {Concat}\left(\text{head}_{1}, \cdots, \text{head}_{H}\right) W^{O},
    \end{array}
    \end{equation}
    where $W_{i}^{Q}$, $W_{i}^{K}$, $W_{i}^{V}$, $W^{O}$ are trainable parameters, and the $\operatorname {Concat}$ means the concatenation operation on the matrix.
    
    To adopt the transformer in the time series domain, the following questions need to be answered~\cite{wen2022transformers}: (1) How to effectively encode the timestamp information into the positional encoding? Since the vanilla transformer treats the input permutation equivalently. (2) How to reduce the complexity with long-range time series as input? Since the vanilla transformer has $O(L^2)$ time complexity ($L$ is the input sequence length).
    Various modified transformers have been proposed to tackle these challenges. For question 1, Lim et al. introduce a trainable positional encoding learned by an LSTM network, which can better extract the sequential information in time series~\cite{lim2021temporal}. Zhou et al. add timestamps as an extra positional encoding to introduce the ordering information to transformer\cite{zhou2021informer}. For question 2, Liu et al.~\cite{liu2021pyraformer} have successfully reduced the time and memory complexity from $O(L^2)$ to $O(L)$, by introducing the sparse attention mechanism~\cite{wen2022transformers}.

    \subsection{GNN-based Models}
    
    Graph neural networks (GNNs)~\cite{kipf2016semi,velivckovic2017graph} have been proven effective in handling graph structure data. This is because GNNs have the ability to capture spatial dependency hidden in non-Euclidean graph structures. This ability can also be used to tackle challenges in modeling multivariate time series,  where GNNs can exploit the latent spatial dependencies between multivariate time series, which the previous models do not fully exploit. Given multivariate time series $\mathcal{X}=\left\{\mathbf{X}_{t_{1}}, \mathbf{X}_{t_{2}}, \cdots, \mathbf{X}_{t_{P}}\right\}$, where $\mathbf{X}_{t_{i}} \in \mathbf{R}^{N \times D}$ denotes the multivariate feature matrix captured at timestamp $i$, and $D$ denotes the feature dimension. A general form of GNN modeling multivariate time series can be denoted as~\cite{wu2020connecting, dai2022graph}:
    \begin{equation}
    \begin{array}{cl}
        \mathbf{A}_{t_{i}} = f_{Map}(\mathbf{X}_{t_{i}}),\\
        \mathbf{z}_{t_{i}} = \operatorname{GNN}(\mathbf{X}_{t_{i}},\mathbf{A}_{t_{i}}),\\
        \mathbf{H} = f_{Sequence}(\mathbf{z}_{t_{1}},..., \mathbf{z}_{t_{P}}),
    \end{array}
    \end{equation}  
    where $f_{map}$ is an adjacency matrix generating function that connects time series to a virtual edge using the information from raw time series data, and $\mathbf{A}_{t_{i}}$ is the generated adjacency matrix at timestamp $i$. GNN denotes a general GNN function, where $\mathbf{z}_{t_{i}}$ is the output of the GNN function. $\mathbf{R}$ is the final result obtained by a sequence modeling function $f_{Sequence}$ using $\mathbf{z}_{t_{1}},..., \mathbf{z}_{t_{P}}$ as input. $\mathbf{H}$ can be applied to various tasks, including forecasting, prediction, and anomaly detection.


    \section{Post-hoc interpretation methods}
    
    We first focus on post-hoc interpretation methods for time series models, which take a pre-trained model of interest and extract interpretable information from it, while keeping the model itself fixed~\cite{du2019techniques}. We will discuss the motivations and methodologies of each category, and analyze their advantages and limitations.
    
    \subsection{Backpropagation-based Methods}
    
\begin{figure*}[htbp]
        \centering
        \includegraphics[width=\linewidth]{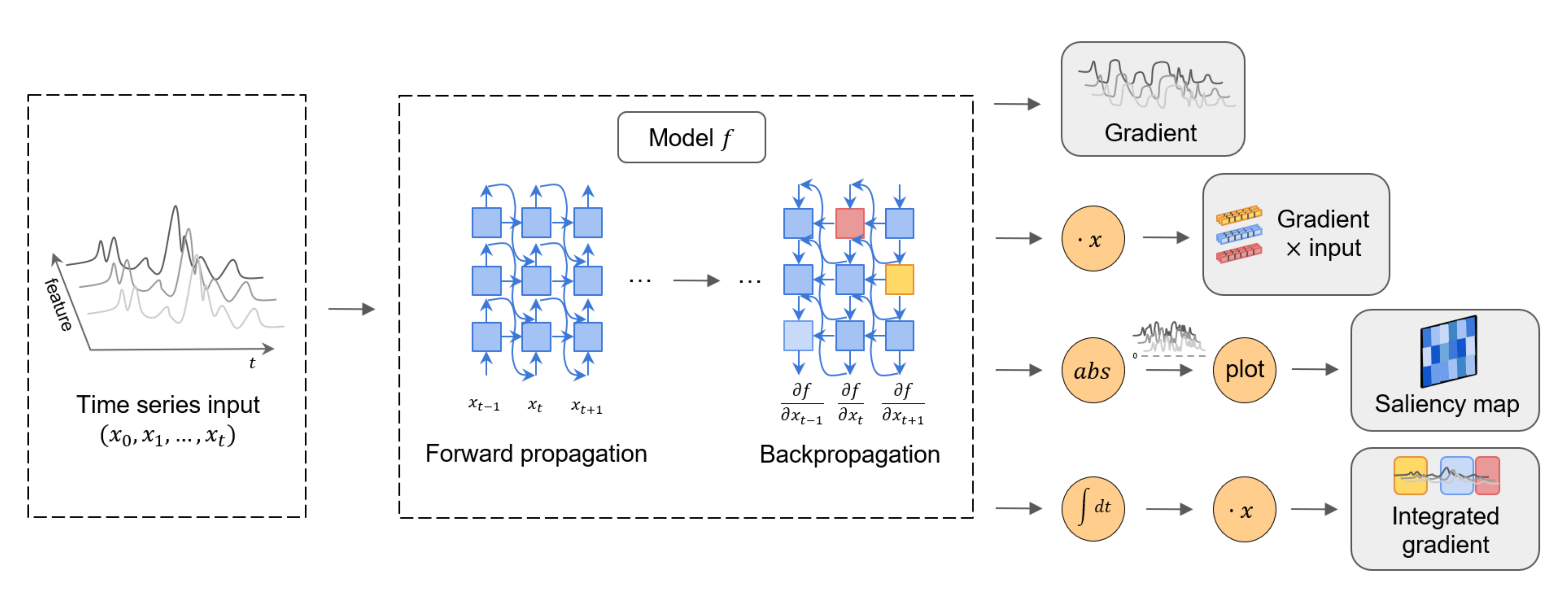}
        \vspace{-15pt}
        \caption{Gradient backpropagation methods for post-hoc interpretation.}
        \label{fig:gradient}
        \vspace{-5pt}
    \end{figure*}

    Backpropagation-based methods operate on a given neural network model by performing backward passes to calculate relevant information that can explain the model's predictions. In this section, we will walk through different types of information that can be backpropagated to help with interpretability. A summary of interpretation methods based on gradient backpropagation is shown in Fig.~\ref{fig:gradient}.

    \subsubsection{Gradient-based}
    
    The most common and fundamental backpropagation methods involve the computation of gradients, i.e. partial derivatives of the neural network~\cite{hechtlinger2016interpretation}. Given a prediction model $f: X\rightarrow Y$ where $X$ is the input space and $Y$ is the output space, and suppose all input features $x\in X$ have the form $x=(x_1, x_2,\dotsc , x_T)$, one can compute the partial derivative
    \begin{equation}
        \nabla f(x) = \left(\frac{\partial f}{\partial x_1}, \frac{\partial f}{\partial x_2},\dotsc , \frac{\partial f}{\partial x_T}\right)
    \end{equation}
    where the resulting vector can be studied to explain the influence of the corresponding variable on the entire model. The importance of each feature can be defined as $I(x_t; f)=\left| \frac{\partial f}{\partial x_t} \right|$. Furthermore, by adapting the mathematical formulation to the exact architecture of the network -- fully-connected, CNN, or RNN, such gradient derivations provide an efficient and intuitive way to analyze the model's sensitivity to different input features. As an example, the Class Activation Map (CAM) method relies on gradient backpropagation in CNNs to find the most contributing region in the original time series data for predicting certain labels~\cite{wang2017time, ismail2019accurate, oviedo2019fast}.
    
    Building upon the basic gradient-based interpretation above, several advanced methods have been proposed to revise the gradient computation to provide more insights~\cite{rojat2021explainable}. For example, the ``gradient$\times$input'' method~\cite{shrikumar2016not, siddiqui2019tsviz} formulates the explanation as the element-wise product of the gradients and the original inputs. Specifically, if we denote the gradient vector as $\nabla f(x) = (\nabla f_1, \nabla f_2,\dotsc, \nabla f_T)$, such method leverages relevant information from the inputs by setting
    \begin{equation}
        I(x_t; f) = x_t\cdot \frac{\partial f}{\partial x_t} .
    \end{equation}
    The above interpretation can be understood as computing the attribution of each feature to the output.
    In addition, the feature attribution can be computed in a more fine-grained manner, by accumulating the attribution along a path between the original input and a baseline point. 
    Such a method, named ``integrated gradient''~\cite{sundararajan2017axiomatic}, can be expressed by setting
    \begin{equation}
        I(x_t; f) = (x_t - \bar{x}_t)\cdot \int_{\alpha=0}^{1}{\frac{\partial f(x')}{\partial x_t} d\alpha}
    \end{equation}
    where $\bar{x} = (\bar{x}_1, \bar{x}_2,\dotsc, \bar{x}_T)$ is the baseline point, and $x' = \alpha x + (1-\alpha) \bar{x}$ represents weighted counterfactual samples. Here $\bar{x}$ is usually set as an all-zero input or the mean value of input samples~\cite{ancona2017towards}. Thus, gradient$\times$input can be regarded as a special case of the integrated gradient by only considering $\alpha=1$.
    
    Gradient-based methods have been widely applied for interpreting healthcare related time series models. For example, the gradient$\times$input method has been adopted to detect myocardial infarction on ECG datasets~\cite{strodthoff2019detecting}. The Clustered Pattern of Highly Activated Period (CPHAP) model, based on gradient$\times$input, is developed to interpret deep temporal representations from human gestures collected by accelerometers and smartphone sensors~\cite{cho2020interpretation}. On the other hand, the ``compensated integrated gradient'' method has been employed to produce reliable explanations for brain activity classification based on EEG datasets~\cite{tachikawa2018compensated}.
    

    \subsubsection{Importance-based}
    
    Another type of information that can be propagated through the network is the importance score. For example, the Layer-wise Relevance Propagation (LRP) method represents feature importance by a ``relevance score'' of an input sample against each prediction class~\cite{bach2015pixel, arras2017explaining}. 
    Given an target prediction $f(x)$ to be interpreted, LRP propagates the prediction score backwards through the network using weighted linear connections, and other techniques such as normalization and pooling.
    The propagation follows the conservation principle that the total amount of relevance distributed to one layer must be equal to that in the previous layer. 
    That is, let $R^{(l)}_i$ denote the relevance score of neuron $i$ at layer $l$, then we have
    $
        \sum_i R^{(l)}_i = f(x) .
    $
    Once the relevance scores are propagated to the input layer where $l=0$, we obtain the contribution, $R^{(0)}_t$, of each input feature $x_t$ to the prediction.
    The LRP method has been adopted in the analysis of machine faults, which relies on high-dimensional temporal data to understand the performance degradation process of machines~\cite{grezmak2019interpretable}, and in the classification of Alzheimer's disease based on structural MRI data~\cite{bohle2019layer}.
    
    Alternatively, another method called DeepLIFT works by comparing each sample against a set of reference inputs and outputs, and explaining the difference between the results~\cite{shrikumar2017learning}. The contribution of each neuron towards the ``difference-from-reference'' is derived by a single backward pass through the network, where the actual signal being propagated reflects the importance of each feature. We can express DeepLIFT by the formula:
    \begin{equation}
        R_i = (x_i - \bar{x}_i)\cdot \frac{\partial^g f}{\partial x_i}
    \end{equation}
    where $\bar{x}$ denotes the baseline sample defined by users, and the partial derivative function $g(x)=\frac{f(x)-f(\bar{x})}{x - \bar{x}}$~\cite{ancona2017towards}.
    Similar to LRP, DeepLIFT also provides several advantages over gradient-based methods, the most crucial one being the added robustness towards zero gradient values and discontinuities caused by bias terms. DeepLIFT delivers comparable performance as LRP in time series interpretation tasks, and they both excel at the explanation of time-series models due to their emphasis on local feature importance~\cite{schlegel2019towards}.

    \subsection{Perturbation-based Methods}
    
    Perturbation-based interpretation methods compute the importance of features by removing, masking, or altering the input, forwarding through the model with the new input, and comparing the change in output predictions~\cite{rojat2021explainable}. Based on the technique of transforming the data, we divide our discussion into the erasure, ablation, and permutation of input features.

    \subsubsection{Erasure}
    
    The most straightforward way of perturbation is through feature erasure, where different parts of the input representation are erased/occluded, and the corresponding effect on the output is observed. This type of method was first introduced in computer vision~\cite{zeiler2014visualizing}, and later revised for the interpretation of time series data. 
    By occluding parts of the input data, ConvTimeNet~\cite{DBLP:journals/corr/abs-1904-12546} used the occlusion sensitivity method to identify the most relevant regions in the time series for making a particular decision. Specifically, the probability of predicting a certain class is defined as $\hat{y}$, and assuming a moving window of size $0.1 \times T$ in time where the input values are set to 0, it follows that whenever an important part of the time series is erased, a sharp drop in $\hat{y}$ is expected. That is, the occlusion sensitivity
    \begin{equation}
        s_t = \hat{y}_t^e - \hat{y}_t^o
    \end{equation}
    will be large, where $\hat{y}_t^e$ is the probability of predicted class after erasure and $\hat{y}_t^o$ is original probability without any modification to input.
    
    In addition to removing input features, one can also erase cells within hidden layers to study the output difference~\cite{li2016understanding}. With a similar evaluation of importance $I(t)$ for dimension $t$ of a selected layer, we have
    \begin{equation}
        I(t) = \dfrac{1}{|X|}\sum_{x\in X}\dfrac{\hat{y}_x - \hat{y}_{(x, \neg t)}}{\hat{y}_x}
    \end{equation}
    where $x \in X$ is a training sample, $\hat{y}_x$ is the probability of correct prediction for $x$ and $\hat{y}_{(x,\neg t)}$ is when dimension $t$ is erased. From such analysis, it was found that at higher layers of a neural model, information importance was spread among different cells within each layer and $I(t)$ was generally lower than that at the input layer.
    Last but not least, feature erasure can not only be used to analyze which parts of the input contribute positively to the final decision, but also function as an error analysis where the removal of certain representation instead improves the model prediction.


    \subsubsection{Ablation}
    
    Feature ablation works by removing $k\geq 1$ features at a time to find the set of features that can most effectively affect model prediction~\cite{ismail2020benchmarking}.  The intuition is simple: when the removal of a certain set of features causes a large difference in the prediction results or their relative confidence, those features are what the model heavily relies on. For example, in the task of determining clinical intervention for ICU patients, ablation helps overcome the noise, sparsity, heterogeneity, and imbalance of the data sources to find meaningful explanations of certain model behaviors~\cite{suresh2017clinical}.
    Another method to characterize the influence of top features is to compute the Ablation Percentage Threshold (APT)~\cite{ozyegen2022evaluation}. Suppose a model $f$ makes predictions based on a total of $T$ features sorted by their importance, and suppose we are ablating $k$ out of these $T$ features. Define
    \begin{equation}
        \text{APT}_{f,\alpha} = \underset{1\leq k\leq T}{\arg\min}\ \frac{k}{T}
    \end{equation}
    with the choice of $k$ satisfying
    \begin{equation}
        f(x_{0,0})(1+\alpha) > f(x_{0,k}) .
    \end{equation}
    Here $x_{0,0}$ denotes the input sample with all features present, $x_{0,k}$ denotes the sample with $k$ ablated features, and $\alpha$ is a hyper-parameter indicating how significant the prediction should change with respect to the removal of features.

    \subsubsection{Permutation}
    
    Instead of completely filtering out features to compute their importance, we can also explain model decisions based on permuted vectors of input variables~\cite{fisher2019all}. 
    The method of Permutation Importance (PIMP)~\cite{altmann2010permutation}, for example, permutes the response vector that accesses feature relevance up to a total of $s$ times, hence generating a vector of $s$ importance scores for each feature. A user-chosen probability distribution is then fitted to the set of all importance vectors, which can be used to derive the probability of observing a feature relevance greater than a certain threshold. PIMP is a generic algorithm for any model that generates feature importance measures, and can perform quite efficiently by appropriate choices of $s$.
    
    Another crucial category of interpretation methods that permutes the input is based on counterfactual or adversarial examples. Unlike previous methods that forward propagating the modified input to observe the output difference, these methods aim at altering the model output to a predefined result (e.g., change the prediction class) and passing this information backwards to generate a new input, thus obtaining explanations about how the model reacts towards input features~\cite{tonekaboni2020explaining, delaney2021instance}. During this process, optimization problems are defined for adversarial perturbation, which usually involve minimizing certain predefined loss functions~\cite{goodfellow2014explaining, liu2018adversarial}.
    
    The loss functions to be considered during the optimization steps depend on the context of the model to be explained. However, one generic form of such functions, as highlighted by the Counterfactual Multivariate Time-series Explainability (CoMTE) method, can be defined as follows~\cite{ates2021counterfactual}. Let $f_c(x)$ denote the probability of the model $f$ predicting the input sample $x$ as output class $c$, and suppose there are $n$ features in the input $x$. We want to find an adversarial sample $x_{adv}$ that leads to a prediction other than $c$, and explain the influence of each feature that takes different values in $x$ and $x_{adv}$. The loss function can then be defined as
    \begin{equation}
        \mathcal{L}(f, c, A, x) = (1 - f_c(x'))^2 + \lambda ||A||
    \end{equation}
    with
    \begin{equation}
        x' = Ax + (I_n - A)x_{adv}
    \end{equation}
    where $A$ is an $n\times n$ diagonal matrix with $A_{jj}=1$ if and only if the $j$-th feature is swapped between $x$ and $x_{adv}$, $||A||$ denotes the matrix rank of $A$ (how many features are swapped in total), and $I_n$ is the $n\times n$ identity matrix. The scalar $\lambda$ is a hyper-parameter to be learned during the optimization steps. The above formulation allows optimal adversarial samples to be discovered, and approximation algorithms to be developed as well to further speed up the explanation process.
    
    Application wise, adversarial training has been employed in the construction of Intrusion Detection Systems (IDS) of network data~\cite{hartl2020explainability}, where the inputs are flows of network packets ordered in a time sequence. By studying how adversarial samples are generated to alter model outputs, one can quantify the feature sensitivity of how likely a feature can lead to mis-classification, and significantly improve the robustness of the model through the obtained explanations. In fact, meaningful conclusions have been drawn from the computation of Adversarial Robustness Score (ARS), such as the crucial vulnerabilities of the first packets, and the importance of characteristic fields such as network protocol and destination port.

    \subsection{Approximation-based Methods}
    
    This type of interpretation methodology is to approximate the feature attribution using local linear models in a model-agnostic manner. The two most representative methods in this category are Local Interpretable Model-agnostic Explanations (LIME) and Shapley Additive Explanation (SHAP), where local feature importance of certain inputs are indicated by the approximated linear coefficients. 

    
    \subsubsection{LIME}
    
    LIME~\cite{ribeiro2016should} explains individual predictions of a black-box model with interpretable local surrogate models, i.e., it interprets the model decision at a specific point by finding out what would happen to the predictions when variations are given to the input. Formally, the model to be explained is defined as $f$, and the class of interpretable models is denoted as $G$, where any $g \in G$ can be readily presented with visual or textual artifacts. LIME first generates a new dataset consisting of perturbed samples around the input data instance $x$ and their corresponding predictions from $f$. Let $\pi_x(z)$ be the proximity measure between an instance $z$ to $x$. On this new dataset, an interpretable model $g$ is trained with the samples weighted according to $\pi_x$. The learned model will hence serve as an approximation to the complex model $f$ locally. The problem of training $g$ is formulated as:
    \begin{equation}
        \label{lime}
        \underset{g \in G}{\arg\min}\ \mathcal{L}(f, g, \pi_x) + \Omega(g)
    \end{equation}
    where $\mathcal{L}(f,g,\pi_x)$ measures the approximation error between $g$ and $f$, and $\Omega(g)$ indicates the complexity of $g$.
    
    Some adaptation should be considered when applying LIME to time-series models~\cite{sivill2022limesegment}. First, it is important to find an interpretable representation of time-series data. The most straightforward way would be to use a pre-determined fixed length window to segment the data \cite{guilleme2019agnostic, neves2021interpretable}, which, however, may suffer from conflicting properties within a segment or homogeneous regions spanning multiple segments due to its arbitrariness. Change point detection is able to solve this problem in some cases by detecting modifications in the statistical properties of a signal~\cite{garreau2018consistent}. Another type of solution focuses on exploring the shape of sub-sequences within the time-series data~\cite{gharghabi2017matrix, zhu2018time}.
    
    The second major difference when dealing with time-series cases would be how to generate samples near $x$ after meaningfully segmenting the data. Existing works involve replacing segments with randomly selected ones from the original dataset~\cite{guilleme2019agnostic}, with mean valued segments~\cite{neves2021interpretable}, or by applying generative perturbations to the time-series data~\cite{sivill2022limesegment}.
    
    The last adaptation to consider is the definition of locality. The traditional Euclidean distance between samples fails to take into account the notion of global distance between generated $x^{(i)}$ and original instance $x$, where multiple $x^{(i)}$ equidistant from $x$ can have different similarities to $x$ based on human cognition. Dynamic Time Warping~\cite{sivill2022limesegment} addressed this issue by ignoring global and local shifts in the time dimension to offer more accurate locality measurement.
    
    Despite the additional efforts required for time-series cases, LIME is a model-agnostic interpretation method that can provide insights for time-series predictions.
    
    \subsubsection{SHAP}

    
    Shapley Additive Explanation (SHAP)~\cite{lundberg2017unified} is another important model-agnostic method to approximate feature attributions. Conceptually, SHAP computes the importance score of a feature by comparing its effect on model prediction when it is present versus absent, across all possible coalitions of other features. It follows the general form of additive feature attribution methods, which explain the model as a linear function $g$ of binary variables
    \begin{equation}\label{eq:shap}
        g(x') = \phi_0 + \sum_{i=1}^{n} \phi_i x_i',
    \end{equation}
    where the binary vector $x'=(x_1',\dotsc,x_n')$ follows either $x_i'=0$ or $x_i'=1$ for each input variable, and values of $\phi_i$ are coefficients denoting the individual attributions.
    
    
    
    
    The SHAP algorithm is computationally expensive since it requires considering all possible combinations of features. To accelerate computation, KernelSHAP has been proposed to approximate the Shapley values. The method is based on the same idea as LIME (see Eq. \ref{lime}), in which it can be proven~\cite{lundberg2017unified} that the only choices of $\Omega$, $\pi_x$ and $\mathcal{L}$ that honor all above three desired properties are the following
    \begin{align}
        \Omega(g) & = 0 \\
        \pi_x(x') & = \frac{n - 1}{\mathbf{C}_n^{|x'|}\, |x'|(n - |x'|)} \\
        L(f, g, \pi_x) & = \sum_{x\in X} \left[f(h_x(x')) - g(x')\right]^2 \pi_x(x')
    \end{align}
    where $\mathbf{C}$ denotes the combination of choices, $|x'|$ refers to the number of active features in $x'$, and $h_x$ is the reverse mapping from binary vector $x'$ to the original sample $x$, i.e. $h_x(x') = x$.

    Adapting SHAP-based methods to time series cases may require some modifications. First, solid proofs have been given that the optimal conditions given in the KernelSHAP method can be fully preserved in the time series domain~\cite{villani2022feature}. The only requirement is to change the linear model $g$ in Eq. \ref{eq:shap} to a Vector Auto-Regressive (VAR) model instead, and the modified method is referred to as VARSHAP. In addition, another method named TimeSHAP has been proposed to properly consider the recurrent nature of time-series data based on KernelSHAP~\cite{bento2021timeshap}. TimeSHAP works by treating each time step as another feature, alongside the set of original features to be studied. Essentially, if we have $n$ features and $d$ time steps in a sequential input $x\in \mathbb{R}^{n\times d}$, then each \textit{row} and \textit{column} of $x$ can both be referred to as a feature.

    One key concern when applying KernelSHAP to time-series models is the exponential computational complexity of SHAP values that arises in the time domain. To address this, the authors of~\cite{villani2022feature} proposed a technique called Time Consistency SHAP values, which formulates a \textit{subgame} for SHAP by reducing the number of time intervals to consider. Specifically, by fixing the observed Shapley values for selected intervals and masking any remaining intervals, the method is able to reduce the computation complexity from $O(2^{nd})$ down to $O(d\times 2^n)$. Alternatively, the TimeSHAP method includes a a temporal coalition pruning algorithm~\cite{bento2021timeshap} that splits the sequence into sub-sequences and computes true Shapley values for each one of them. The resulting complexity is $O(d\times 2^{n(d-i)})$, where $i$ denotes the number of pruned time steps after the algorithm. Both methods considerably improve the scalability of KernelSHAP and make the computation affordable for larger time-series models.

    The KernelSHAP method and its variants have been employed to explain model behaviors in multiple problem domains, including the usage analysis of household electronics~\cite{villani2022feature} and account takeover fraud detection~\cite{bento2021timeshap}. Another work~\cite{mokhtari2019interpreting} also extends the SHAP method to the analysis of financial time-series data to predict the commentaries learned from financial experts' reports.

    \begin{table*}[ht]
    \begin{minipage}{\linewidth}
        \centering
        \caption{Summary of interpretation methods for time series models and tasks.}
        \label{method summary}
        \vspace{-10pt}
        \renewcommand{\arraystretch}{1.2}
        \renewcommand{\thempfootnote}{\fnsymbol{mpfootnote}}
        \begin{tabular}[t]{lcccc}
            \hline
            Model & Methodology & Post-hoc/Ante-hoc & Problem Types\footnote[2]{C = Classification, D = Detection, F = Forecasting, G = Generation, R = Regression 
            } & Model Agnostic/Specific\\
            \hline\hline
            Basic gradient interpretation~\cite{hechtlinger2016interpretation} & Backpropagation & Post-hoc & C, R & Specific\\
            CAM~\cite{wang2017time, ismail2019accurate, oviedo2019fast} & Backpropagation & Post-hoc & C & Specific\\
            Gradient $\times$ input~\cite{shrikumar2016not, siddiqui2019tsviz, strodthoff2019detecting} & Backpropagation & Post-hoc & C, D, F, R & Specific\\
            CPHAP~\cite{cho2020interpretation} & Backpropagation & Post-hoc & D & Specific\\
            Integrated gradients~\cite{sundararajan2017axiomatic} & Backpropagation & Post-hoc & C, D, F & Specific\\
            Compensated integrated gradients~\cite{tachikawa2018compensated} & Backpropagation & Post-hoc & C & Specific\\
            Saliency map~\cite{simonyan2013deep, assaf2019mtex} & Backpropagation & Post-hoc & C, F, R & Specific\\
            LRP~\cite{arras2017explaining, grezmak2019interpretable, bohle2019layer} & Backpropagation & Post-hoc & C, F & Specific\\
            DeepLIFT~\cite{shrikumar2017learning} & Backpropagation & Post-hoc & C, F & Specific\\
            Erasure~\cite{li2016understanding} & Perturbation & Post-hoc & C, D, F & Specific\\
            ConvTimeNet~\cite{DBLP:journals/corr/abs-1904-12546} & Perturbation & Post-hoc & C & Specific\\
            LSTM feature-level occlusion~\cite{suresh2017clinical} & Perturbation & Post-hoc & F & Specific\\
            APT~\cite{ozyegen2022evaluation} & Perturbation & Post-hoc & F & Agnostic\\
            PIMP~\cite{altmann2010permutation} & Perturbation & Post-hoc & R & Agnostic\\
            Counterfactual and adversarial samples~\cite{tonekaboni2020explaining, delaney2021instance, goodfellow2014explaining, liu2018adversarial} & Perturbation & Post-hoc & C, D, G, R & Agnostic\\
            CoMTE~\cite{ates2021counterfactual} & Perturbation & Post-hoc & C, D, R & Specific\\
            ARS~\cite{hartl2020explainability} & Perturbation & Post-hoc & G & Specific\\
            LIMEsegment~\cite{sivill2022limesegment} & Approximation & Post-hoc & C & Agnostic\\
            Temporal attention mechanism~\cite{karim2017lstm, schockaert2020attention, choi2019prediction, ge2018interpretable} & Attention & Ante-hoc & F, G, R & Specific\\
            Spatial-temporal attention~\cite{gangopadhyay2021spatiotemporal, lin2020preserving} & Attention & Ante-hoc & F & Specific\\
            Temporal contextual layer~\cite{vinayavekhin2018focusing} & Attention & Ante-hoc & C, D 
            & Specific\\
            RETAIN~\cite{choi2016retain} & Attention & Ante-hoc & F & Specific\\
            Self-attention~\cite{huang2019dsanet, garnot2020satellite, du2023saits} & Attention & Ante-hoc & C, F, G & Specific\\
            DTS~\cite{li2021learning} & Disentangle rep. & Ante-hoc & G & Specific\\
            IMGC~\cite{marcinkevivcs2021interpretable} & Causality & Ante-hoc & R & Specific\\
            CAP~\cite{dhaou2021causal} & Causality & Ante-hoc & C & Specific\\
            TCDF~\cite{nauta2019causal} & Causality & Ante-hoc & F & Specific\\
            ADSN~\cite{ma2020adversarial}& Shapelets & Ante-hoc & C & Agnostic\\
            ShapeNet~\cite{li2021shapenet}& Shapelets & Ante-hoc & C & Specific\\
            Physics-based convolution~\cite{nascimento2020tutorial,sadoughi2019physics,wehmeyer2018time,wu2022novel} & Physics rule & Ante-hoc & F & Specific\\
            Physics-based regularization~\cite{raissi2019physics,zhu2019physics, geneva2020modeling, kissas2020machine} & Physics rule & Ante-hoc & F, R & Agnostic\\
            \hline
        \end{tabular}
    \end{minipage}
    \end{table*}

    \section{Inherently interpretable models}\label{sec:inherently}
    
    \subsection{Attention Mechanism}

    
    One popular approach that could extend neural networks for better interpretability is the attention mechanism~\cite{chaudhari2021attentive}. At its core, the method computes a weighted context vector as a conditional distribution over input sequences, and the layer that learns these weights is often embedded in the model structure~\cite{rojat2021explainable}. The key assumption to any attention-based method is that the assigned weights correspond to the relative importance of different input segments. 

    \subsubsection{Basic Formulation}
    
    We use RNNs to illustrate the fundamental idea of attention mechanism. Suppose an RNN is trained with input length of $n$ and output length of $m$, and suppose the network encodes the input at step $i$ using an embedding $h_i$, with $i = 1,\dotsc,n$. Let $c_t$ denote the output vector at time step $t$. Then, $c_t$ for each output position $t=1,\dotsc,m$ is defined as a linear combination of all the input embeddings, which can be computed by
    \begin{equation}
        c_t = \sum_{i=1}^n \alpha_{t,i}h_i .
    \end{equation}
    Here $\alpha_{t,i}$ is the \textit{attention weights} after applying a softmax function over the scores $e_{t,i}$, where
    \begin{equation}
        \alpha_{t,i} = \frac{\exp(e_{t,i})}{\sum_{j=1}^n \exp(e_{t,j})} ,
    \end{equation}
    with each weight $e_{t,i}$ being a learnable function $a$ over input embedding $h_i$ and output state $s_{t}$ of the RNN on time step $t$, i.e.,
    $
        e_{t,i} = a(s_{t}, h_i) .
    $
    The attention weight $\alpha_{t,i}$ or $e_{t,i}$ is a measure of how much the input at position $i$ contributes to the output at position $t$. In order to parameterize such weights, one could train a feedforward neural network jointly with the RNN, so the explanation is available immediately during model inference~\cite{bahdanau2014neural}.

    The time-series attention mechanism has been utilized in various application including ironmaking industry~\cite{schockaert2020attention}, UCR dataset analysis~\cite{karim2017lstm}, temporal medical image analysis~\cite{choi2019prediction}, and ICU mortality prediction~\cite{ge2018interpretable} to provide interpretability to their models.
    In order to account for the correlation between spatial and temporal information, the concept of ``spatial-temporal attention''~\cite{gangopadhyay2021spatiotemporal, lin2020preserving} is introduced to simultaneously learn the most important time steps and feature variables.
    Furthermore,~\cite{vinayavekhin2018focusing} proposes the ``temporal contextual layer'', which calculates attention $e_{t,i}$ for each time step $t$ based on information from the whole input sequence, i.e., \begin{equation}
        e_{t,i} = a(h_1, h_2,\dotsc, h_n).
    \end{equation}
    and results in more focused attention values and more plausible visualization than traditional methods. Finally, for some time-series data best studied in retrospect, the RETAIN (Reverse-Time Attention) model~\cite{choi2016retain} is proposed for the interpretation of EHR data, whose most important information lies in the most recent clinical visits. 
    
    
    
    \subsubsection{Self-attention}
    Different from the traditional attention mechanism where explanations are made based on RNNs or CNNs, self-attention explores the relationship of elements within a sequence entirely on its own~\cite{vaswani2017attention, devlin2018bert}. It is at the heart of Transformers, where each element in the sequence attends to every other one and their relationship is captured thereby. Specifically, assume the input consists of queries and keys of dimension $d_k$, and the queries, keys and values are packed into $Q, K$ and $V$ respectively. Self-attention is therefore computed as
    \begin{equation}
        A = \text{softmax}(\dfrac{QK^T}{\sqrt{d_k}})V
    \end{equation}
    
    The result is calculated as the weighted sum of $V$, where the weights of each position is the inner product between $Q$ and $K$ at every other position in the time series~\cite{huang2019dsanet}. To counteract the effect that when $d_k$ is very large, the softmax function will be pushed into regions where gradients are extremely small, and the attention value was scaled by a factor of $\dfrac{1}{\sqrt{d_k}}$.

    One application in time series domain of self-attention would be to capture the relationships among multiple time series data~\cite{huang2019dsanet}. In the paper, the authors proposed a dual self-attention network (DSANet) which utilized two parallel convolution components to capture temporal patterns and a self-attention module to model the dependencies among multiple sequences. Besides, self-attention can also be employed to extract temporal features within one time series~\cite{garnot2020satellite}, where it can usually outperform RNNs on a variety of tasks. Moreover, a model named Self-Attention-based Imputation for Time Series (SAITS) was developed which shows that self-attention can be used to fill in missing values in a time series~\cite{du2023saits}.
    
    Self-attention has the advantage of lowering computational complexity per layer and it increases parallelism compared to RNNs. It also requires less depth when learning features within long sequences which makes it more desirable over CNNs.

\subsubsection{Controversy}

Despite the attention mechanism being a straightforward interpretation method, there are controversies around whether it provides meaningful insights for model predictions or not. The opponents believe that the attention weights from original models only weakly correlate with the feature importance derived by conventional gradient-based and ablation-based methods~\cite{jain2019attention}. Their major argument comes from the fact that adversarial samples with unique attention weights can still be constructed leading to equivalent predictions, which indicates that attention mechanisms may not be faithful to the grounded interpretations of model behaviors. Meanwhile, the proponents advocates that attention weights are capable of demonstrating meaningful interpretation, and the adversarial aspects of attention mechanisms does not necessarily hold due to the ambiguous definition on model explanation~\cite{wiegreffe2019attention}. Currently, the interpretability issue of model attention weights is still being debated across the whole community, but a conclusion we can safely draw is that attention weights can somewhat reflect the model behaviors under certain circumstances.

\subsection{Interpretable Representation Space Design}
Learning effective representations is the core of modern deep learning models. Instead of extracting understandable information from representations in post-hoc manners, it is also possible to proactively design interpretable representation space. In this part, we discuss several typical paradigms for learning interpretable representations from data.

\subsubsection{Disentangled Representation Learning}

Disentangled representation learning refers to the process of learning representations of data that separate the underlying factors of variation~\cite{locatello2019challenging}. In the context of time series data, disentangled representation learning involves separating sources of variation that contribute to different sequences or segments. For example, in financial time series data, the underlying factors of variation could include market trends, seasonal patterns, and individual company performance~\cite{cao2003support, anvari2016disentangling}. Formally speaking, the goal of disentanglement is to learn a set of decomposed latent variables ${z_1,z_2,...,z_N}$, where any two variables $z_i$ and $z_j$ are independent when $i \neq j$. Such representations have latent dimensions containing diverse and mutually independent semantic information~\cite{li2021learning}. We can interpret the latent factor contained in each dimension as an explanation.


There are two primary approaches to disentangled representation learning for time series data: generative-based methods and contrastive-based methods. Generative-based methods focus on learning a low-dimensional representation of the data by modeling the generative process of the time series, while contrastive-based methods aim to identify the most informative features by contrasting pairs of samples with respect to a similarity metric. In the following sections, we provide an overview of each approach, including its underlying principles and relevant examples.

Disentangled generative-based methods on time series data usually involve training an autoencoder model to reconstruct the original data while learning a disentangled latent representation. To encourage disentanglement, regularization terms like the evidence lower bound (ELBO)~\cite{kingma2013auto} are added to the loss function. For instance, Li et al.~\cite{li2021learning} apply LSTM-VAE to capture temporal correlations in time series data and disentangle learned representations by adding a penalty parameter to the vanilla VAE reconstruction loss. Another example is the factorizing variational autoencoder (FAVAE) proposed by Yamada et al.~\cite{yamada2020disentangled}. This method employs the information bottleneck principle to learn disentangled representations from sequential data using a variational autoencoder as the backbone. The information bottleneck principle restricts the flow of information through a network, encouraging it to focus on the most important features of the data. Li et al.~\cite{li2022generative} further combine disentangled learning with diffusion and denoise processes and proposed D3VE. In addition, latent variables in D3VE are disentangled on top of minimizing total correlation to enhance the interpretability and stability of the prediction. 

Disentangled contrastive-based methods on time series data aim to differentiate between pairs of time series that are either similar (positive pairs) or different (negative pairs). The learned representations are expected to capture underlying independent factors that distinguish the pairs. For example, Woo et al.~\cite{woo2022cost} propose CoST, that utilizes contrastive learning to learn disentangled seasonal-trend representations via time domain and frequency domain contrastive losses. Chen et al.~\cite{chen2021learning} introduced DeepDGL, a deep forecasting model with an encoder-decoder architecture that disentangles dynamics into global and local temporal patterns. Their model employs vector quantization and a contrastive multi-horizon coding based adaptive parameter generation module to model diversified and heterogeneous patterns. In addition, Gowda et al.~\cite{gowda2022contrastive} applied a contrastive and trend-seasonal disentangled representation learning approach to generate synthetic time series sequences for battery datasets, addressing the challenge of acquiring high-quality battery datasets. 

\subsubsection{Shapelet Learning}
Time series models with shapelets have attracted considerable research interest, due to their high discriminative ability and good interpretability. 
Shapelets refer to the time series sub-sequences that can effectively discriminate between different time series types~\cite{ye2009time,grabocka2014learning}. 
Shapelets offer effective time series primitives to enhance model interpretability. For instance, we can use distances between time series and shapelets as features for downstream tasks, such as classification, where feature space more understandable compared to using raw time points~\cite{lines2012shapelet}.

Various methods have been developed to discover shapelets, such as extracting shapelets from a pool of candidates in the input time series~\cite{ye2009time,ye2011time}, and learning shapelets by optimizing specifically designed objective functions~\cite{grabocka2014learning}.
Recently, some researchers have employed adversarial training to overcome the trade-off dilemma between performance and interpretability by learning shapelets that resemble subsequences of the input time series~\cite{wang2019learning, ma2020adversarial}. 
In addition, ShapeNets~\cite{li2021shapenet} learns shapelets with different lengths from multivariate time series by embedding the variable-length shapelets into a unified space through minimization of the triplet loss function and then selecting the top-k representative shapelets as the final shapelets. Those methods of learning shapelets are able to balance the interpretability and performance of deep learning models. 

\subsubsection{Symbolic Representation Learning}
Symbolic representation learning incorporates Symbolic Aggregate Approximation (SAX) and~\cite{lin2003symbolic,lin2007experiencing} in the time domain and Symbolic Fourier Approximation (SFX)~\cite{schafer2015boss,schafer2016scalable} in the frequency domain to interpretable time series models involves transforming raw time series data or frequency domain features into symbolic representations that capture important patterns and relationships in a more interpretable format.  
The process of SAX consists of several steps. 

First, the raw time series data is preprocessed to clean and normalize it. Then, symbolization is performed by partitioning the time series into segments and assigning symbols to each segment based on specific criteria. Symbolization methods can include amplitude-based, threshold-based, or entropy-based approaches. Next, feature extraction techniques are applied to the symbolic representations to capture relevant characteristics. These features can include statistical measures, frequency analysis, or complexity measures. Once the features are extracted, interpretable time series models are built using machine learning or statistical techniques. Examples of such models are decision trees \cite{senin2013sax,yan2022neuro}, rule-based models, symbolic regression, or fuzzy logic models. These models explicitly represent rules or patterns that humans can easily understand and interpret. 

Different from the SAX, SFA use Discrete Fourier Transformation (DFT) to discretize the time series. SAX maps the time series data into a predefined symbolic alphabet, while SFA retains the original continuous values in terms of Fourier coefficients. is Fourier coefficients. SAX is primarily concerned with capturing the shape and distribution of data, while SFA focuses on identifying dominant frequency components. A combination of multi-resolution (i.e. multiple SAX representations) and multi-domain symbolic representations (SAX and SFA representations) enables more robust results in applications~\cite{le2019interpretable}.

\subsection{Causality-based Models}

Causality represents the relation between a cause and its effect.
Humans generate explanations based on causal beliefs and reasoning, i.e., identifying the relationship between a cause and its effect~\cite{danks2009psychology}. Therefore, a model becomes more explainable if it considers causality between input and output~\cite{kaddour2022causal,moraffah2020causal}. In this section, we introduce the causality-based models on time series data.

\subsubsection{Causal Interpretability}

Causality-based models are interpretable compared to black-box models, since they are designed to explicitly represent the causal relationships between variables ~\cite{beckers2022causal}. Such explicit makes it easier to interpret the model input and output relationships in a clear and transparent manner.

Causal inference and causal discovery are two main topics concerning causality. Causal inference seeks to estimate the causal effect of one variable on another, while causal discovery seeks to discover the underlying causal structure of a system ~\cite{zanga2022survey}. In this survey, we will focus on causal discovery for inherent interpretability, since our aim is to find cause-effect pairs as part of model interpretation.

By using causal discovery techniques, we can analyze the relationship between input variables and a model's predictions and identify which variables are significant causes of predicted outcomes. For example, current studies~\cite{imtiaz2013correlation} have found that blood pressure values are correlated to the relative position of peaks in the seismocardiogram (SCG) curve.  If a causal discovery analysis reveals that the peak parts are a significant factor in the model's blood pressure predictions, we can have greater confidence in the model's output and understand why it made those predictions.

In the following subsections, we will first discuss a representative causality discovery method: Granger causality. We then introduce two other popular and interpretable causality methods: constraint-based methods and score-based methods.

\subsubsection{Granger Causality}

Granger causality (GC) is one of the first proposed causal interactions in multivariate time series modeling~\cite{granger1969investigating}. The approach has been applied to various fields, such as neuroscience~\cite{roebroeck2005mapping}, economics~\cite{appiah2018investigating}, and climatology~\cite{charakopoulos2018dynamics}. GC aims to disclose the past values of a time series in order to predict the future values of another series. The original GC is based on the linear model (i.e., vector autoregressive model (VAR)). Let $\mathbf{x}_t$ denote a time series vector at time $t$, where $\mathbf{x}_t = (x_{1t},x_{2t},...,x_{pt})^T$, then GC performed on a VAR model can be defined as follows~\cite{shojaie2022granger}:
\begin{equation}
        A^{0} \mathbf{x}_{t}=\sum_{k=1}^{L} A^{k} \mathbf{x}_{t-k}+\mathbf{e}_{t}
\end{equation}
where L denotes the time lags, and $A_0, A_1,..., A_L$ are lag matrices in $p \times p$ dimensions. The $\mathbf{e}_{t}$ is a Gaussian random vector with $0$ mean, which is the white noise~\cite{shojaie2022granger}. 

GC is a powerful tool for analyzing multivariate time series data. However, it is only reliable on stationary time series. When dealing with non-stationary data, it may lead to wrong causal inference~\cite{tian2001causal}.Recent efforts have been made to overcome this limitation and discover causality in non-stationary time series using approaches such as surrogate variables~\cite{zhang2017causal}.
 
Additionally, various approaches related to GC have been proposed to understand and interpret time series models. For example,  Hu and Liang~\cite{hu2014copula} propose a copula-based model-free GC measure that can reveal nonlinear, higher-order causality, overcoming the problem that traditional GC cannot capture high-order causal effects. Their variability assessment strategy is based on resampling and allows for statistical significance tests for derived high-order GC. Marcinkevic et al.~\cite{marcinkevivcs2021interpretable} propose a novel framework to infer the GC of multivariate time series in nonlinear dynamics. This framework is interpretable because it can not only infer the relation, but also detect effects signs and inspect their variability over time with the help of self-explaining neural networks (SENNs). 


\subsubsection{Constraint-based Interpretable Causality Models}

Constraint-based models rely on a set of causal assumptions and use statistical tests to identify causal relationships consistent with the assumptions~\cite{spirtes2000causation}. These models seek to find a causal structure that best fits the data given the assumptions, and as a result, they require strong prior assumptions about the causal relationships among variables, such as acyclicity (no cycles in the causal graph), causal sufficiency (every variable has a direct cause), and faithfulness (the statistical dependencies in the data match the causal relationships).

For instance, Dhaou et al.~\cite {dhaou2021causal} present a framework for understanding the causes of phenomena that occur briefly over time, such as flooding. Their framework includes the Case-crossover APriori (CAP) algorithm, which provides association and causal rules that explain the occurrences of these phenomena, and the Case-crossover APriori Predictive algorithm (CAPP1 and CAPP2) that predicts these causal rules. Chattopadhyay et al.~\cite{chattopadhyay2019neural} propose that neural network architectures can be considered as structural causal models, and present a method to compute the causal effect of each feature on the output. With reasonable assumptions on the causal structure of the input data, they propose causal regressors that estimate the causal effects efficiently. Their effectiveness is proven on the RNNs and compared with the gradient-based explanation method (IG).

\subsubsection{Score-based Interpretable Causality Models}

One popular category of causality discovery techniques is the score-based method~\cite{bengio2019meta}, which involves selecting a score function and searching through the space of possible causal relationships for the optimal ones that minimize this score function. Score-based methods usually do not rely on prior assumptions about causal relationships among variables.

For example, Pamfil et al.~\cite{pamfil2020dynotears} introduce DYNOTEARS, a method for discovering time-lagged and contemporaneous relationships among variables in a time series. The approach involves minimizing a penalized loss while utilizing an acyclicity constraint that is characterized as a smooth equality constraint. Mansouri et al.~\cite{mansouri2020heidegger} propose the HEIDEGGER framework for interpretable temporal causal discovery, which consists of a flexible randomized block design and an efficient causal profile search that has been tested on the cognitive health dataset to find out which lifestyle factors matter at reducing cognitive decline. In addition,  Nauta et al. propose a deep learning framework called Temporal Causal Discovery Framework (TCDF) for both time series prediction and temporal causal discovery, which consists of multiple layers of CNN with attention mechanisms. Their framework can identify causal relationships hidden within the time series data, identify the time delay between each cause and effect, and construct a causal graph based on the causality with delays~\cite{nauta2019causal}.

\subsection{Physics Rule-based Models}
Physics-guided deep learning integrates physics laws with hard-to-interpret deep learning models by guiding deep models to a path following physical rules. In this way, the physics-guided deep learning models can be inherently interpretable~\cite{karniadakis2021physics}. The physics laws can provide prior information or constraints on the model space, thereby increasing the model's robustness to noisy data~\cite{karniadakis2021physics, rai2020driven}. Efforts to combine the physics laws and machine learning models can be divided into two main categories, one is the physics-based architecture design, and the other is physics-based regularization.  
Physics-based architecture approaches embed the physics knowledge into the neural network architecture, while the physics-based regularization adds a penalized term to the loss function to constrain model parameter space.
\begin{figure*}[htbp]
\centering
\includegraphics[width=0.8\linewidth]{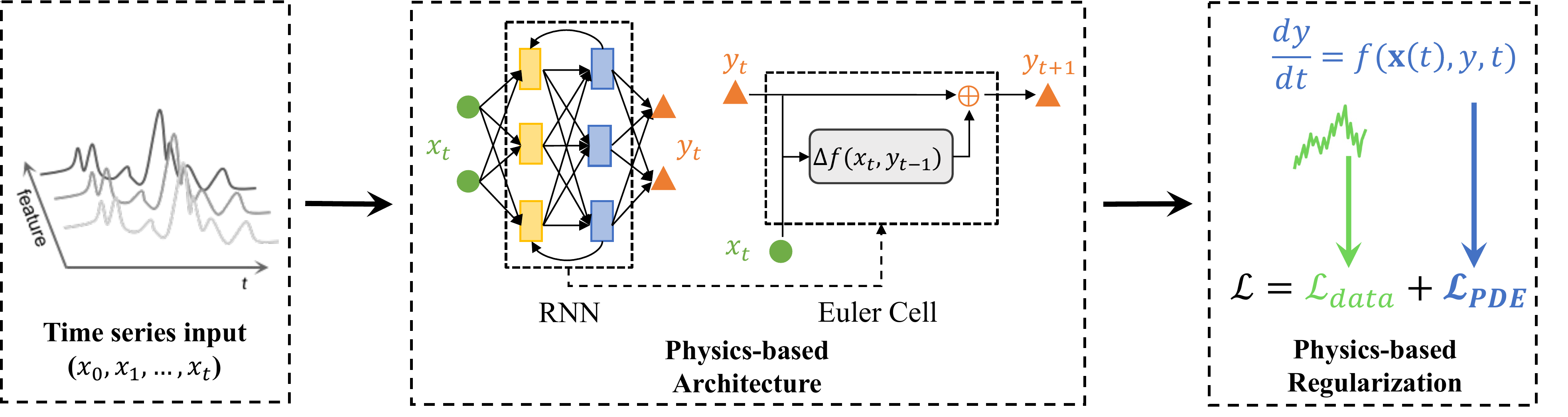}
\caption{An illustration of physics rule-based models.}
\vspace{-10pt}
\end{figure*}

\subsubsection{Physics-based Architecture}
The lack of interpretable information and integration of the existing physical laws in the real-world motivated some researchers to propose a novel neural network architecture to integrate the underlying physical laws of the system.
Partial differential equations (PDEs) is a powerful mathematical language to express spatial-temporal interaction in multivariate time series. The most common way of integrating the existing physical knowledge is to incorporate PDEs in the machine learning architecture and guarantee the model prediction satisfies the given PDEs~\cite{nascimento2020tutorial, zhang2020physics,darbon2021some}.

To be more specific, we illustrate how to design novel cells assimilating the LSTM and GRU cells in a Recurrent Neural network, i.e. a cell for Euler integration~\cite{butcher2016numerical, nascimento2020tutorial}. The so-called Euler cell can implement numerical integration while training the neural network, instead of going through gated in LSTM and GRU.

The first-order ordinary differential equation can be expressed by:
\begin{equation}\label{eq:ode}
\frac{d y}{d t}=f(\mathbf{x}(t), y, t),
\end{equation}
where $\mathbf{x}(t)$ are the inputs to the dynamic system, $y$ is output to be predicted, and $t$ is times. The solution to the above equation depends on the initial condition and input. To solve the differential equation, we can simply use Euler's method in the neural network,
\begin{equation}\label{eq:euler}
y_n=y_0+\sum_{t=1}^{n}f(x_t,y_{t-1},\mathbf{w},\mathbf{b}),
\end{equation}
where $\mathbf{w}$ and $\mathbf{b}$ are unknown parameters to be estimated in the neural network to approximate the unknown function $f(t)$.
To solve Equation~\ref{eq:ode}, we can use the observed data to minimize the loss function after:
\begin{equation}
L=\frac{1}{n}\sum_{t=1}^{n}\left[y_t-(y_0+\hat{f}(x_t,y_{t-1},\mathbf{w},\mathbf{b}))\right]^2,
\end{equation}
where $\hat{f}$ is the prediction function with input $x_t$ at the current time point, in replace of the original LSTM or GRU cell in the RNN model.
Thus, the time series going through the Euler cells in the updated RNN is doing numerical integration. Such cells mimicking various integration techniques, i.e., Runge–Kutta integration~\cite{butcher2016numerical} can be designed in RNN models to solve the differential equations and forecast.

In addition to specific PDEs, there are also some generalized physical knowledge that can be applied~\cite{karniadakis2021physics}, such as the property of the dynamic system~\cite{sadoughi2019physics}, general physical laws~\cite{udrescu2020ai} embedded into the neural network by developing novel processing schemes, designing novel convolutions, and specific basis functions tailored for the physics-based architecture modeling. 
For example, ~\cite{ling2016reynolds} designs a novel architecture with an invariant tensor basis to embed Galilean invariance property for better predictions of the turbulence models. ~\cite{sadoughi2019physics} uses physics knowledge in bearings to generate reference kernels and design physics-based convolutions. ~\cite{wehmeyer2018time} develops a time-lagged autoencoder that maps the latent vector to the lagged decoder to fit molecular dynamics. ~\cite{wu2022novel} develops a novel multi-resolution convolutional interaction network to capture the temporal dependencies at multiple resolutions to improve forecasting accuracy.
Instead of changing the kernels,~\cite{udrescu2020ai} develops an AI-Feynman algorithm that integrates dimensional analysis, polynomial fit, and neural networks to discover the governing functions of the data.

\subsubsection{Physics-based Regularization}
In addition to designing novel architectures for the neural network, another line of research focuses on introducing learning biases by the physics knowledge in deep learning models. Physics-informed neural network (PINN)~\cite{raissi2019physics} is a typical example that integrates PDEs into the loss function of a neural network. Assuming that we have a PDE following the equation 
\begin{equation}
\frac{\partial y}{\partial t}+y \frac{\partial y}{\partial x}=v \frac{\partial^2 y}{\partial x^2},
\end{equation}
and measurements of $u(x,t)$ collected at observed spatial-temporal points. We have the following loss function to train the neural network:
\begin{equation}
\mathcal{L}=w_{\text {data }} \mathcal{L}_{\text {data }}+w_{\text {PDE }} \mathcal{L}_{\text {PDE }},
\end{equation}
where
\begin{equation}
\begin{aligned}
& \mathcal{L}_{\text {data }}=\frac{1}{N_{\text {data }}} \sum_{i=1}^{N_{\text {data }}}\left(y\left(x_i, t_i\right)-y_i\right)^2 \,\, , \\
& \mathcal{L}_{\text {PDE }}=\left.\frac{1}{N_{\mathrm{PDE}}} \sum_{j=1}^{N_{\text {PDE }}}\left(\frac{\partial y}{\partial t}+y \frac{\partial y}{\partial x}-v \frac{\partial^2 y}{\partial x^2}\right)^2\right|_{\left(x_j, t_j\right)},
\end{aligned}
\end{equation}
where $v$ is the input diffusion coefficient, $w_{\text {data }}$ and $w_{\text {PDE }}$ are tuning parameters that balance the importance of loss terms $\mathcal{L}_{\text {data }}$ and $\mathcal{L}_{\text {PDE }}$ from fitting the data and physical laws, respectively.

In addition to the PINN, there are other variants~\cite{zhu2019physics, geneva2020modeling, kissas2020machine} that constrain the solution by physical laws and allow interpretable uncertainty quantification. ~\cite{zhu2019physics} use an encoder-decoder to solve a PDE with constraint, ~\cite{geneva2020modeling} solve the conservation laws in graph topology in an arterial network, and ~\cite{kissas2020machine} employed a deep auto-regressive model to solve non-linear PDE and provided a Bayesian framework for uncertainty quantification. Physics laws are served as constraints in loss function while training the deep learning models for time series data, thus, the partial differential equations are parameterized by convolutional blocks, and forecasting and regression tasks are easy to handle.

    \section{Interpretation Evaluation}
    
    As more methods for interpreting time series data are developed, the need increases for evaluation methods and metrics to assess their effectiveness in explaining the prediction process and results. This helps to avoid blindly trusting interpretation results without empirical evidence of their reliability. 

    \subsection{Qualitative Evaluation}\label{sec:qual_eval}
    
    There are mainly two types of evaluation metrics that are commonly used for time-series interpretation methods. The first category relies on domain knowledge, where experts are involved to determine the relevancy of explanations. Since the evaluation of interpretation in some medical fields is hard to quantify, bringing in clinical expert insights and experience would be important~\cite{ho2021interpreting}. Visualization methods can also be used to empirically evaluate whether the explanation presented matches established principles in a specific field~\cite{kim2019electric}. However, the employment of such evaluations in time-series interpretation can be very limited, due to the difficulty of intuitively understanding time-series data~\cite{schlegel2019towards}. Therefore, it would be more accurate to use metrics for quantitative-based evaluation, as described in the remaining subsection.

    \subsection{Quantitative Evaluation}
    
    \subsubsection{Perturbation Analysis}
    
    The method of perturbation analysis in the image domain could be adapted to evaluate time series interpretations~\cite{schlegel2019towards}. The evaluation pipeline on time series models borrows the idea of setting relevant pixels to zero in images, and changes the time series sequence in a similar fashion to determine the correctness of explanations. Specifically, we denote a time series $t$ and its relevance score vector $r$ as
    \begin{equation}
        t = (t_1, t_2,\dotsc, t_n),\ r = (r_1, r_2,\dotsc, r_n) .
    \end{equation}
    We could choose to focus on either time points or sequences in evaluation.
    In the first case, perturbation analysis selects a threshold $e$, such as the 90th percentile of values in $r$, and set all $t_i=0$ or $t_i = \max(t) - t_i$ for any $r_i > e$. The perturbed samples are denoted as $t^{zero}$ or $t^{inverse}$, respectively. 
    In the second case, one could perturb certain sub-sequences of the time series instead of individual points, by the \textbf{swap time points} method where the entire sub-sequence of length $n_s$ after the first $r_i > e$ is reversed
    \begin{equation}
        (t_i, t_{i+1}, t_{i+2}, \dotsc, t_{i+n_s}) \rightarrow (t_{i+n_s}, \dotsc, t_{i+2}, t_{i+1}, t_i) ,
    \end{equation}
    or the \textbf{mean time points} method where the mean value $\mu_{seq}$ of the same sub-sequence substitutes all original values
    \begin{equation}
        (t_i, t_{i+1}, \dotsc, t_{i+n_s}) \rightarrow (\mu_{seq}, \mu_{seq}, \dotsc, \mu_{seq}),
    \end{equation}
     to take into account the inter-dependency of time points. The perturbed samples would then be predicted by the model to compute the quality metric, such as the decrease in accuracy, to quantify how well the interpretation method has been performing.

    Another type of method that uses perturbation to evaluate the interpretation of regression tasks is \textbf{area over the perturbation curve for regression (AOPCR)}~\cite{ozyegen2022evaluation}. In this method, the features were sorted according to the local explanation metric, where the top $K$ features were excluded from $\mathbf{X}_t$, defined as $\mathbf{X}_{t,\setminus 1:K}$. The AOPCR at time $\tau$ is obtained as
    \begin{equation}
        \text{AOPCR}_\tau = \dfrac{1}{K} \sum_{k=1}^K f_\tau (\mathbf{X}_t) - f_\tau (\mathbf{X}_{t, \setminus 1:K}) .
    \end{equation}
    The total AOPCR is the average from all time steps $1, \dots t_0$
    \begin{equation}
        \text{AOPCR} = \dfrac{1}{t_0} \sum_{\tau = 1}^{t_0} \text{AOPCR}_\tau .
    \end{equation}
    This evaluation metric can be applied to time-series prediction problems.

    \subsubsection{Orthogonal Metrics}
    
    Additionally, a framework of six orthogonal metrics was proposed~\cite{loffler2022don} to assess the quality of post-hoc interpretation methods for time-series classification and segmentation. The six metrics -- namely the \textbf{faithfulness} of model outputs w.r.t. input features, the \textbf{robustness} of the model against adversarial attacks, the \textbf{sanity} of model weights and biases, the \textbf{localization} of high-relevance features within the latent space, the \textbf{inter-class sensitivity} and the \textbf{intra-class sensitivity} of explanations -- could all be computed based on standard assumptions on the interpreted time series model. Though the usage of such framework is limited to post-hoc explanations only, its assessment on model interpretation quality is proven to be simple, clear and effective.

    \subsubsection{Precision and Recall}
    
    Precision and recall were used in another study to determine the explanation quality~\cite{ismail2020benchmarking} of saliency methods, where identified salient features were examined by the weighted precision and recall of each (\textit{neural architecture, saliency method}) pair. Precision indicated whether all identified features were informative, and recall suggested whether all informative features were included. Specific metrics used in the evaluation were area under the precision curve (AUP), area under the recall curve (AUR), and area under precision and recall (AUPR), where the curves were calculated at different levels of degradation. According to the evaluations, the study found that commonly-used saliency methods failed to produce high-quality interpretation when applied to multivariate time-series.

    \section{Future directions and challenges}

    In this section, we will briefly go over some current challenges we recognized in the field of time-series interpretation, and bring forward some research directions that may serve as a guideline for future studies.

    First and foremost, the interpretation of time series models should serve the purpose to explain their behaviors to \textbf{non-experts} in machine learning or data science, in order to be trusted in crucial use cases such as medical treatments. While all work presented here appeals to the formality of an explanation, many of the results are still far away from being intuitive enough for users with non-technical backgrounds. For example, gradient-based explanations stem from a few key concepts in calculus, while the importance scores assigned by methods like KernelSHAP require knowledge of combinatorics to fully understand. It is urgent that we call for \textbf{more intuitive explanations} to be devised for the studies of any relevant models.

    Secondly, current explanations are rarely used to \textbf{improve upon existing models}. In most situations, explanations are only developed for understanding the model predictions and exploring important features. While this has been the primary usage, explanation results can also be used to detect flaws in architecture design or remove redundant features~\cite{li2016understanding}. It would be very helpful to have more researches focus on how to use existing explanation methods to analyze the model and features being interpreted and make corresponding adjustments upon them.

    Thirdly, while a few efforts have been made to distinguish between certain categories of interpretation methods, such as~\cite{schlegel2019towards} for importance versus approximation-based methods, and~\cite{ancona2017towards} for different versions of backpropagation, the \textbf{comparison} of different means of explanation has been lacking in general. There are quite a few important questions yet to be understood, such as the \textbf{advantages and drawbacks} of inherently interpretable models as opposed to post-hoc methods, and how each novel method presented in Section \ref{sec:inherently} performs and compares to each other in an analysis of similar time-series data.
    
    And finally, as mentioned in Section \ref{sec:qual_eval}, for evaluation of model explanations, \textbf{quantitative evaluation} is preferred over \textbf{qualitative evaluation} due to its more accurate nature. However, in most cases, evaluating a model interpretation involves significant amount of domain knowledge, where qualitative metrics are usually easier to use. In fact, a lot of researches have evaluated their models qualitatively due to this appeal of being more straightforward, but they missed out the opportunity to characterize their explanations using precise numbers and definitive evidence. Therefore, it is important that more quantitative metrics with \textbf{domain-specific context} are developed in order to draw more accurate conclusions.

    \section{Conclusion}


    Our survey paper focuses on the interpretability of time-series data, providing a comprehensive overview of research directions and techniques. We discuss commonly employed models for time-series classification and forecasting tasks, including CNN, RNN, transformer, and GNN. Additionally, we comprehensively explain the mechanisms of post-hoc interpretation methods, such as backpropagation-based, perturbation-based, and approximation-based approaches, evaluating their strengths and limitations. Furthermore, we introduce inherently interpretable models, such as attention-based models, interpretable representation learning models, causality-based models, and physics rule-based models, which offer inherent transparency without the need for additional post-hoc interpretation methods. We also present a range of evaluation metrics to quantify time-series model interpretability and highlight existing challenges while proposing potential research directions.

    

    \bibliographystyle{ieeetr}
    \bibliography{ref.bib}

\end{document}